\documentclass[letterpaper]{article} % DO NOT CHANGE THIS
\usepackage{aaai2026}  % DO NOT CHANGE THIS
\usepackage{times}  % DO NOT CHANGE THIS
\usepackage{helvet}  % DO NOT CHANGE THIS
\usepackage{courier}  % DO NOT CHANGE THIS
\usepackage[hyphens]{url}  % DO NOT CHANGE THIS
\usepackage{graphicx} % DO NOT CHANGE THIS
\usepackage{natbib}  % DO NOT CHANGE THIS AND DO NOT ADD ANY OPTIONS TO IT
\usepackage{caption} % DO NOT CHANGE THIS AND DO NOT ADD ANY OPTIONS TO IT
\usepackage{algorithm}
\usepackage{algorithmic}
\usepackage{bm}
\usepackage{amsmath,amssymb,amsfonts}
\usepackage{multirow}
\usepackage{makecell}

\usepackage{newfloat}
\usepackage{listings}
\DeclareCaptionStyle{ruled}{labelfont=normalfont,labelsep=colon,strut=off} % DO NOT CHANGE THIS
\floatstyle{ruled}
\newfloat{listing}{tb}{lst}{}
\floatname{listing}{Listing}
\title{Generating Sketches in a Hierarchical Auto-Regressive Process for Flexible Sketch Drawing Manipulation at Stroke-Level}
\author{
    Sicong Zang\thanks{Corresponding author.}, Shuhui Gao, Zhijun Fang\\
}
\affiliations{
    School of Information and Intelligent Science, Donghua University, Shanghai 201620, China\\
    sczang@dhu.edu.cn, 2242994@mail.dhu.edu.cn, zjfang@dhu.edu.cn
}

\usepackage{bibentry}
\begin{document}

\maketitle

\begin{abstract}
Generating sketches with specific patterns as expected, i.e., manipulating sketches in a controllable way, is a popular task. Recent studies control sketch features at stroke-level by editing values of stroke embeddings as conditions. However, in order to provide generator a global view about what a sketch is going to be drawn, all these edited conditions should be collected and fed into generator simultaneously before generation starts, i.e., no further manipulation is allowed during sketch generating process. In order to realize sketch drawing manipulation more flexibly, we propose a hierarchical auto-regressive sketch generating process. Instead of generating an entire sketch at once, each stroke in a sketch is generated in a three-staged hierarchy: 1) predicting a stroke embedding to represent which stroke is going to be drawn, and 2) anchoring the predicted stroke on the canvas, and 3) translating the embedding to a sequence of drawing actions to form the full sketch. Moreover, the stroke prediction, anchoring and translation are proceeded auto-regressively, i.e., both the recently generated strokes and their positions are considered to predict the current one, guiding model to produce an appropriate stroke at a suitable position to benefit the full sketch generation. It is flexible to manipulate stroke-level sketch drawing at any time during generation by adjusting the exposed editable stroke embeddings.
\end{abstract}

% Uncomment the following to link to your code, datasets, an extended version or similar.
% You must keep this block between (not within) the abstract and the main body of the paper.
\begin{links}
    \link{Code}{https://github.com/SCZang/Sketch-HARP}
\end{links}

%%%%%%%%%%%%%%%%%%%%%
\section{Introduction}
%%%%%%%%%%%%%%%%%%%%%

Free-hand sketches are valuable tools to convey messages and express emotions. Generative models are trained to produce sketches with specific patterns as expected, i.e., manipulating sketches in a controllable way. Several applications have been introduced to verify whether a model could manipulate sketches accurately and robustly. Controllable sketch synthesis \cite{zang2021controllable} requires generate sketches to preserve both categorical and stylistic patterns from the input sketches. Sketch healing \cite{su2020sketchhealer} targets to restore missing details from corrupted sketches. Sketch analogy \cite{ha2017neural} aims to generate sketches with hybrid patterns collected from multiple sketches given. These applications are instance-level sketch manipulations, i.e., the adjustments are applied on the full sketch instead of partial components. As a result, it is not flexible enough to precisely control some specific local contents (e.g., adjusting the shape of a single stroke) in a reliable manner.

\begin{figure}[!t]
    \centering
    \includegraphics[width=\columnwidth]{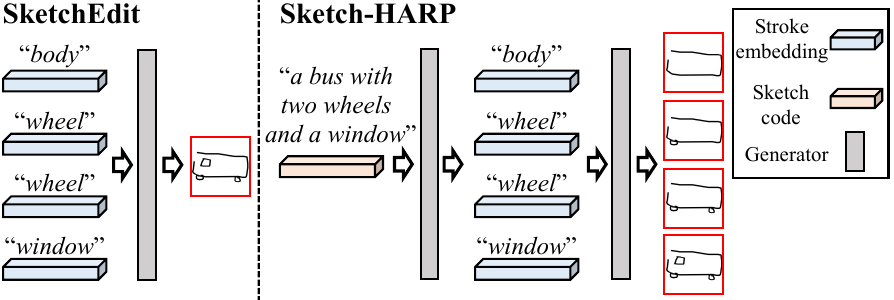}
    \caption{Manipulating sketches at stroke-level by SketchEdit \cite{li2024sketchedit} and the proposed Sketch-HARP. For SketchEdit, the editable stroke embeddings are simultaneously collected and fed into the generator before generation starts, i.e., no manipulation is allowed during sketch generating process. Sketch-HARP generates sketches in a hierarchical process, i.e., with a given sketch code, a group of stroke embeddings are firstly generated, which are secondly translated into drawing actions and positioned on the canvas. The exposed stroke embeddings are editable and could be flexibly adjusted during sketch drawing to enable stroke-level sketch drawing manipulation.}
    \label{fig:comparison}
\end{figure}

Recently, SketchEdit \cite{li2024sketchedit} is proposed to manipulate sketches at stroke-level. Each sketch is separated by a group of strokes with paired 2D coordinates to locate them on the canvas, and it is able to control features and locations of strokes by editing their captured embeddings' values. However, as shown in Fig. \ref{fig:comparison}(left), when manipulating sketches by SketchEdit, in order to provide its generator a global view about what a sketch is going to be drawn, all the adjusted stroke embeddings should be simultaneously collected and fed into its generator before generation starts. As a result, no further manipulation is allowed during sketch generating process.

We propose a hierarchical auto-regressive sketch generating process for flexible sketch drawing manipulation at stroke-level. On one hand, our generative process is in a hierarchical manner. Instead of producing the entire sketch at once, we generate stroke embeddings to bridge the starting sketch code with the final generated sketch, shown in Fig. \ref{fig:comparison}(right). More specifically, each stroke is generated in three stages: firstly predicting a stroke embedding to represent which stroke is going to be drawn, and secondly determining the stroke's position on the canvas, and finally translating stroke embedding to a sequence of drawing actions to form the full sketch. Such a generating process imitates human sketching procedure well: 1) realizing which sketch component is going to be drawn, and 2) considering where to settle the pen, and 3) moving the pen to draw a stroke. Moreover, as stroke embeddings are exposed during sketch generating process, we are able to edit them at any time during generation to realize flexible stroke-level sketch drawing manipulation.

On the other hand, the separated stroke embedding prediction, position determination and embedding translation are all auto-regressively processed. When generating a stroke, the recently produced strokes along with their positions are both considered as references, which requires model to have a global view about what the current unfinished sketch would be like and further to generate an appropriate stroke at a suitable location to benefit sketch generation. 

To realize these above, we propose a method, namely sketch generation in a Hierarchical Auto-Regressive Process (Sketch-HARP) for flexible sketch drawing manipulation at stroke-level. A stroke encoder and a position encoder are employed to capture embeddings for strokes and their starting positions on the canvas, respectively, and a relationship encoder is for incorporating strokes' features before sent into a sketch encoder to obtain sketch codes. The sketch code is fed into a hierarchical auto-regressive generator to recursively produce paired stroke embeddings, starting positions, and drawing actions step-by-step in a queue. To summarize, we make the following contributions:
\begin{itemize}
	\item[1.] We introduce Sketch-HARP to generate sketches in a hierarchical process to enable flexible sketch drawing manipulation at stroke-level.
	\item[2.] We propose an auto-regressive generator to recursively generate sketch strokes and their positions on the canvas by turns, producing appropriate strokes at suitable locations to benefit sketch generation.
	\item[3.] Experimental results indicate that Sketch-HARP achieves accurate and robust performance on many sketch manipulating applications, such as stroke replacement, stroke erasion, stroke expansion, etc.
\end{itemize}

%%%%%%%%%%%%%%%%%%%%%
\section{Related Work}
%%%%%%%%%%%%%%%%%%%%%

\subsection{Sketch Generation}

Learning accurate sketch representations is a common approach to improve sketch generation. Sketches are represented by sequences of drawing actions \cite{ha2017neural,wang2024self,li2024sketchedit}, raster images \cite{chen2017sketch,li2024lmser,li2025sketchmlp}, or both \cite{su2020sketchhealer,zang2025equipping} for learning contextual information among strokes and visual patterns from canvas. Besides, self-organizing sketch representations into specific structures in the latent space could also regularize networks to benefit sketch learning. For example, constructing a Gaussian mixture model (GMM) distributed latent space \cite{zang2021controllable,zang2024self} would encourage sketches with similar patterns to be clustered together, further contributing to accurate and robust sketch pattern representation.

Employing powerful network structures is another effective approach. The long short-term memory (LSTM) \cite{graves2012long} based generator proposed by \cite{ha2017neural} is widely used to produce a sequence of drawing actions to describe sketches \cite{zang2023linking,li2024sketchedit,su2020sketchhealer}. Its generating process imitates how humans draw a sketch stroke-by-stroke. Diffusion model \cite{ho2020denoising} based architectures are also utilized to morph and locate strokes for constructing recognizable sketches \cite{wang2023sketchknitter,das2023chirodiff}.

In this paper, we propose a new sketch generating process by predicting the drawn strokes step-by-step before generating a full sketch, which benefits stroke-level sketch drawing manipulation in a flexible way.

\subsection{Sketch Manipulation}

Sketch manipulation aims to generate sketches with expected patterns in a controllable way. Controllable sketch synthesis \cite{zang2021controllable}, sketch healing \cite{su2020sketchhealer}, sketch reorganization \cite{wang2024self}, sketch analogy \cite{ha2017neural,zang2024self} and sketch edit \cite{li2024sketchedit} are sketch manipulating applications. Recent studies manipulate sketches by adjusting values of sketch codes or stroke embeddings, and the edited features are fed into generators to generate sketches as expected. Their manipulating operations (e.g., editing stroke embeddings) cannot be applied during sketch drawing.

The proposed Sketch-HARP divides sketch generating process into stroke embedding prediction, position determination and embedding translation. We are able to adjust stroke embeddings via replacement, expansion, erasion, etc. during sketch drawing, driving to flexible sketch drawing manipulation at stroke-level. 

%%%%%%%%%%%%%%%%%%%%%
\section{Methodology}
%%%%%%%%%%%%%%%%%%%%%

\begin{figure*}[t]
    \centering
    \includegraphics[width=1.6\columnwidth]{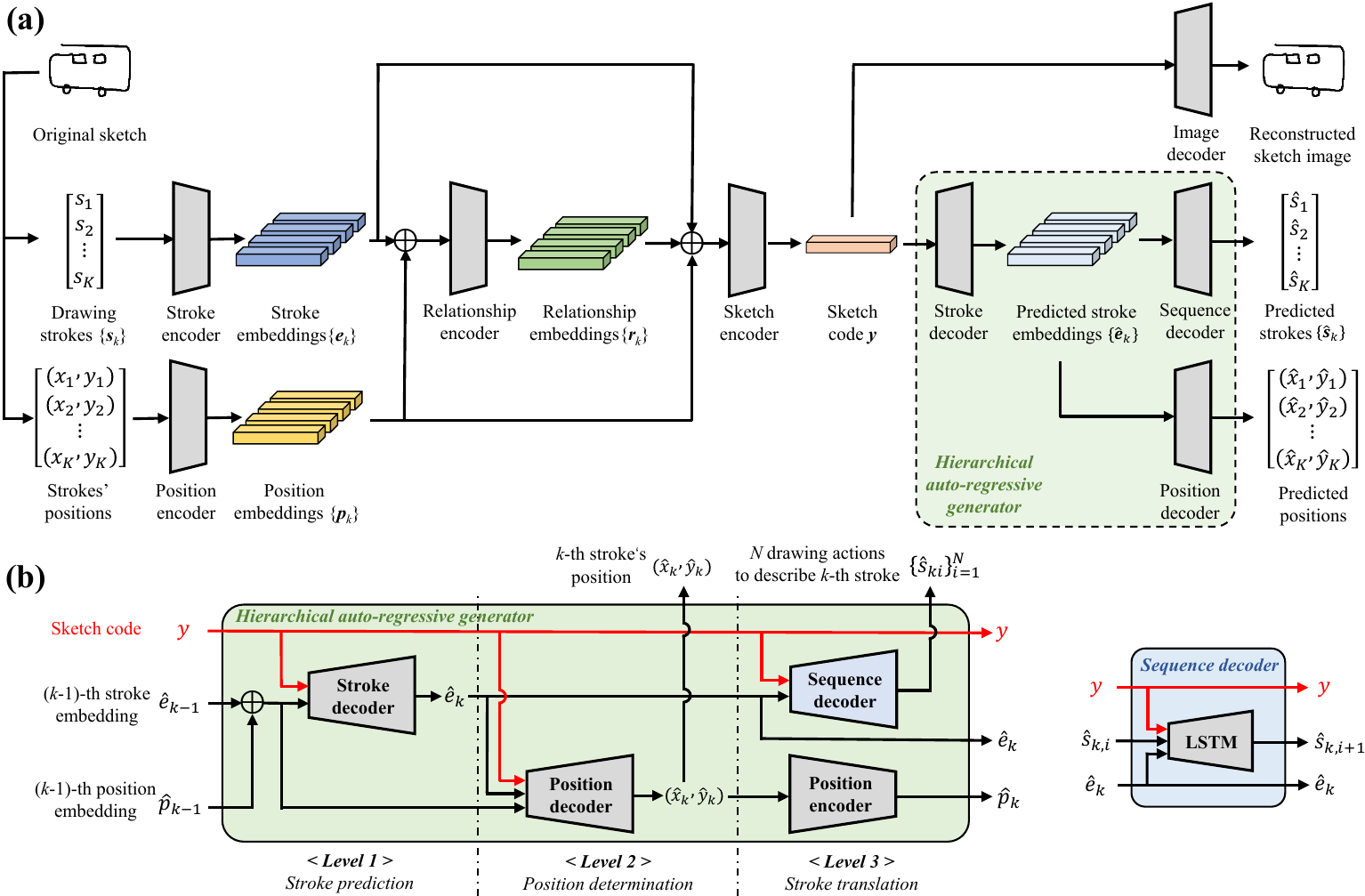}
    \caption{Generating sketches in a hierarchical auto-regressive process by Sketch-HARP. (a) The network structure. We learn relationships among strokes and incorporate them with features from strokes and starting positions to obtain a sketch code, which is fed into a hierarchical auto-regressive generator to produce sketches. (b) Generating sketches in a hierarchical auto-regressive process. Each stroke is generated by three stages: 1) predicting which stroke is going to be drawn, 2) determining where to locate the predicted stroke, and 3) translating stroke embeddings into drawing actions to finally form a full sketch.}
    \label{fig:overview}
\end{figure*}

Fig. \ref{fig:overview} offers the network structure of Sketch-HARP. A sketch is separated into sequence-formed strokes with their starting positions on the canvas, whose captured embeddings are incorporated to obtain a sketch code. Sketch generation is conditional on the learned code in a hierarchical auto-regressive way. The code is sent into a stroke decoder to generate a group of stroke embeddings recursively, which are fed into a sequence decoder and a position decoder to produce their corresponding positions and to translate them into drawing actions to form a full sketch, respectively. 

\subsection{Translating Sketches into Paired Strokes and Starting Positions}

In \emph{QuickDraw} \cite{ha2017neural}, a sequence-formed sketch consists of a series of drawing actions $(x, y, l_{0}, l_{1}, l_{2})$ to record how it is drawn step-by-step. $(x, y)$ is a two-dimensional coordinate to store the position of pen on the canvas, and $(l_{0}, l_{1}, l_{2})$ is a one-hot vector to indicate three pen states (pen down, pen lifting and end-of-drawing). As a stroke ends when $l_{1}=1$, we could split a sketch with $K$ strokes into $K$ groups of drawing actions. The $k$-th stroke $\bm s_k$ ($k=1, 2, \cdots, K$) collects $j$ drawing actions $\{\bm s_{k1}, \bm s_{k2}, \cdots, \bm s_{kj}\}$, as it takes $j$ steps to finish its drawing. Besides, we mark the 2D coordinate of the first point of $\bm s_k$ as $(x_k, y_k)$, namely the starting position of stroke $\bm s_k$. With the cooperation between $\bm s_k$ and $(x_k, y_k)$, we can organize strokes on the canvas to construct a full sketch. 

\subsection{Learning Sketch Representations}

For a single stroke $\bm s_k$, we firstly capture its stroke and position embeddings to restore its trajectory of pen moving and its location on the canvas, respectively. In practice, a bidirectional LSTM-based stroke encoder $q_{\text{sok}}$ is adopted to capture stroke embedding $\bm e_k\in\mathbb{R}^{128\times 1}$ from $\bm s_k$. A fully connected layer, working as a position encoder $q_{\text{pos}}$, is employed to project 2D coordinates into position embeddings $\bm p_k\in\mathbb{R}^{128\times 1}$. 
\begin{align}
	\label{eq:stroke_position_embedding}
    \bm e_k = q_{\text{sok}}(\bm s_k), \quad\bm p_k = q_{\text{pos}}((x_k, y_k)).
\end{align}

Such captured embeddings $\bm e_k$ and $\bm p_k$ only describe what and where a stroke is, but could hardly tell which semantic component it would be to construct a full sketch. For example, a tiny round-shaped stroke might be a wheel or a window in a bus. Actually, the semantic meaning of a stroke for constructing a sketch is determined by not only its shape, but also the relationships among strokes. For example, we might realize a tiny circle to be a wheel when we figure out that it locates beneath a large rectangle as the body of a bus. Such relationships between strokes might cover: 1) spatial relationships, indicating the relative positions of strokes on the canvas (e.g., nose is always positioned above mouth in a pig); 2) contextual relationships, reporting the temporal distances between strokes in the drawing order (e.g., two eyes of a pig are drawn closely in the drawing order); and 3) semantic relationships, revealing sketch-component-level dependencies among strokes for constructing a recognizable sketch (e.g., exact two ears should be drawn to complete a pig).

Thus, it is beneficial to equip a stroke embedding with information passing from the other strokes. When realizing what other strokes are by $\{\bm e_k\}$ and where they are located by $\{\bm p_k\}$, Sketch-HARP would have a global view about how strokes are organized in a full sketch, further learning comprehensive stroke representations.

We employ a gMLP block \cite{liu2021pay} as relationship encoder $q_{\text{rel}}$ to learn relationship embedding $\bm r_k\in\mathbb{R}^{128\times 1}$. $\bm r_k$ is further incorporated to $\bm e_k$ to yield the final stroke representation $\tilde{\bm e}_k$.
\begin{align}
	\label{eq:relationship_embedding}
    \tilde{\bm e}_k = \bm e_k + \bm r_k,\quad\bm r_k = q_{\text{rel}}(\{\bm e_k + \bm p_k\}_{k=1}^K).
\end{align}

Finally, we employ an LSTM-based sketch encoder $q_{\text{skc}}$ to obtain the final sketch code $\bm y\in\mathbb{R}^{128\times 1}$.
\begin{align}
	\label{eq:learn_sketch_code}
    \bm y = q_{\text{skc}}(\{\tilde{\bm e}_k+\bm p_k\}_{k=1}^K).
\end{align}
The sequential information about stroke drawing order would be restored in code $\bm y$.

\subsection{Generating Sketches via a Hierarchical Auto-Regressive Process}

When generating sketches, a sketch code is fed into a series of decoders to produce the predicted stroke embeddings, their starting positions, and further drawing actions recursively in a hierarchical process, shown in Fig. \ref{fig:overview}(b).

\noindent\textbf{(I) Predicting stroke embeddings} 

In the first stage of generation, we utilize an LSTM-based stroke decoder $p_\text{sok}$ to predict stroke embeddings $\{\hat{\bm e}_k\}_{k=1}^K$.
\begin{align}
	\label{eq:predict_stroke_emb}
    \hat{\bm e}_{k}, \hat{\bm\eta}_{k}=p_\text{sok}(\bm y, \tilde{\bm e}_{k-1}+\bm p_{k-1}).
\end{align}
We feed $p_\text{sok}$ with a given sketch code $\bm y$ and the $(k-1)$-th enriched stroke embedding $\tilde{\bm e}_{k-1}$ along with its position embedding $\bm p_{k-1}$ to predict $\hat{\bm e}_k$. The stroke decoder receives information about which sketch is being drawn (from $\bm y$), which stroke has been drawn previously (from $\tilde{\bm e}_{k-1}$) and where it locates (from $\bm p_{k-1}$), and further predicts $\hat{\bm e}_k$ as an appropriate stroke to be drawn currently. Note that a pair of tokens $\tilde{\bm e}_0$ and $\bm p_0$, regarded as starting markers with their values by $\tilde{\bm e}_0=\bm p_0=-\bm 1$, are introduced to predict the first stroke embedding $\hat{\bm e}_1$.

Furthermore, in Eq. (\ref{eq:predict_stroke_emb}), a two-dimensional vector $\hat{\bm \eta}_k=[\hat{\eta}_{k0}, \hat{\eta}_{k1}]$ ($\hat{\eta}_{k0}+\hat{\eta}_{k1}=1$) is also produced simultaneously with the prediction of $\hat{\bm e}_k$. $\hat{\bm \eta}_k$ is a marker to reveal whether the currently predicted stroke would be the last one, i.e., guiding Sketch-HARP to realize when to stop stroke production. $[\hat{\eta}_{k0}, \hat{\eta}_{k1}]$ is set as parameters of a categorical distribution to model the ground truth $\bm\eta_{k}$. $\bm\eta_k=[1, 0]$, if the $k$-th stroke exists, and $k=\min\:\{i\left |\bm\eta_i=[0, 1]\right .\}$ indicates that the $k$-th stroke is the last one. Thus, we could use $\hat{\bm\eta}$ to determine the number of strokes in sketch generation. 

\noindent\textbf{(II) Anchoring strokes on the canvas}

A group of ordered stroke embeddings $\{\hat{\bm e}_k\}_{k=1}^K$ has been predicted, but where to position them on the canvas has not been determined yet. Inspired by \cite{ha2017neural}, we model a probability distribution function (P.D.F.) to predict 2D coordinates as starting positions for each $\hat{\bm e}_k$. More specifically, the P.D.F. is a bi-variate Gaussian distribution $\mathcal{N}\left (x, y|\mu^{px}, \mu^{py}, \sigma^{px}, \sigma^{py}, \rho^p\right )$, where $\mu^{px}$, $\mu^{py}$, $\sigma^{px}$, $\sigma^{py}$ and $\rho^p$ denote the means, the standard deviations and the correlation parameter, respectively. When anchoring a stroke on the canvas, we sample a 2D coordinate from the distribution as the predicted position. In practice, we use an LSTM-based position decoder $p_{\text{pos}}$ to predict the distribution parameters.
\begin{align}
	\label{eq:predict_pos_emb}
    \mu^{px}_{k}, \mu^{py}_{k}, \sigma^{px}_{k}, \sigma^{py}_{k}, \rho^p_{k}=p_{\text{pos}}(\bm y, \hat{\bm e}_{k-1}+\bm p_{k-1}, \hat{\bm e}_k).
\end{align}

When modeling the P.D.F. of the $k$-th stroke's position, the position decoder is always fed with three inputs. The first one is a sketch code $\bm y$, which reminds $p_{\text{pos}}$ what a sketch is currently being drawn. The second one is $\hat{\bm e}_{k-1}+\bm p_{k-1}$, which offers the information about which stroke has already been drawn recently and where it has been located. The final input is $\hat{\bm e}_k$, whose position is being determined currently. Besides, when anchoring the first stroke, we also introduce a token $\hat{\bm e}_0=-\bm 1$ as a starting marker.

\noindent\textbf{(III) Generating sequenced-formed strokes}

We have produced a group of stroke embeddings as representations of strokes, along with their 2D coordinates on the canvas. To finish sketch drawing, the final step is translating these predicted stroke embeddings $\{\hat{\bm e}_k\}$ into sequence-formed strokes by drawing actions. We adopt the LSTM-based sequence decoder $p_{\text{seq}}$ from \cite{ha2017neural} to generate a sequence of drawing actions $\hat{\bm s}_k=\{[\Delta \hat{x}_{ki},\Delta \hat{y}_{ki},\hat{\bm l}_{ki}]\}_{i=1}^{N_k}$ for each stroke, where $N_k$ denotes the length of the $k$-th stroke. More specifically, when generating the $i$-th drawing action for $\hat{\bm s}_k$, $p_{\text{seq}}$ produces: 1) a mixture of $M$ bi-variate Gaussian distributions $\sum_{m=1}^M\alpha_{kim}\cdot\mathcal{N}\left (\Delta x_{ki}, \Delta y_{ki}|\mu^x_{kim}, \mu^y_{kim}, \sigma^x_{kim}, \sigma^y_{kim}, \rho_{kim}\right )$ with mixing weights $\{\alpha_{kim}\}_{m=1}^M$ to model the P.D.F. of offset distance $(\Delta x_{ki}, \Delta y_{ki})$ between the $i$-th and the $(i-1)$-th drawing actions in the $k$-th stroke, and 2) a categorical distribution with parameter $\hat{\bm l}_{ki}$ to predict its pen state for alerting model to stop drawing.
\begin{align}
	\label{eq:predict_action}
    \resizebox{0.9\hsize}{!}{$
    \left \{\bm\alpha_{km}, \bm\mu^{x}_{km}, \bm\mu^{y}_{km}, \bm\sigma^{x}_{km}, \bm\sigma^{y}_{km}, \bm\rho_{km}\right \}_{m=1}^M, \hat{\bm l}_k=p_{\text{seq}}(\bm y, \hat{\bm e}_k).
    $}
\end{align}

In addition, as the relationships among strokes have been modeled into the predicted stroke embeddings $\{\hat{\bm e}_k\}$ via their auto-regressive generating processes, we could decode one stroke embedding without the attendance of other strokes. Hence, it is possible to translate multiple $\hat{\bm e}_k$ simultaneously into drawing actions. 

\begin{algorithm}[t!]
\caption{Generating a sketch from a given sketch code by Sketch-HARP}
\label{algo1}
\textbf{Input}: Sketch code $\bm y$\\
\textbf{Output}: Collection of strokes' drawing actions $\mathcal{\bm S}$, collection of 2D coordinates $\mathcal{\bm P}$
\begin{algorithmic}[1]
\STATE $\hat{\bm e}_0\gets-\bm 1$, $\hat{\bm p}\gets-\bm 1$, $\mathcal{\bm S}\gets\emptyset$, $\mathcal{\bm P}\gets\emptyset$, $k\gets 0$ %\COMMENT{Initialization}
\WHILE{$k<N_{\text{max}}^{\text{num}}$} %\Comment{Max number of strokes}
    \STATE $\hat{\bm e}_1, \hat{\bm\eta}\gets p_{\text{sok}}(\bm y, \hat{\bm e}_0+\hat{\bm p})$ %\Comment{Stroke decoder in Eq. (\ref{eq:predict_stroke_emb})}
    \STATE $\bm\eta_0\sim\text{Categorical}(\hat{\bm\eta})$ %\Comment{Sample a marker}
    \IF{$\bm\eta_0=[0, 1]$} %\Comment{Finish sketch drawing}
        \STATE \textbf{break}
    \ENDIF
    \STATE $\mu^{px}, \mu^{py}, \sigma^{px}, \sigma^{py}, \rho^p\gets p_{\text{pos}}(\bm y, \hat{\bm e}_0+\hat{\bm p},\hat{\bm e}_1)$ %\Comment{Position decoder in Eq. (\ref{eq:predict_pos_emb})}
    \STATE $(\hat{x}, \hat{y})\sim\mathcal{N}(\mu^{px}, \mu^{py}, \sigma^{px}, \sigma^{py}, \rho^p)$ %\Comment{Sample a 2D coordinate}
    \STATE $\mathcal{\bm P}\gets\mathcal{\bm P}\cup\{(\hat{x},\hat{y})\}$ %\Comment{Save stroke's starting position}
    \STATE $\hat{\bm p}\gets q_{\text{pos}}((\hat{x},\hat{y}))$ %\Comment{Learn position embedding}
    \STATE $\hat{\bm e}_0\gets\hat{\bm e}_1$
    \STATE $i\gets 0$, $\mathcal{\bm A}\gets\emptyset$
    \WHILE{$i<N_{\text{max}}^{\text{len}}$} %\Comment{Max length of a stroke}
        \STATE $\bm\alpha,\bm\mu^x,\bm\mu^y,\bm\sigma^x,\bm\sigma^y,\bm\rho,\hat{\bm l}\gets p_{\text{seq}}(\bm y,\hat{\bm e}_1)$ %\Comment{Sequence decoder in Eq. (\ref{eq:predict_action})}
        \STATE $(\Delta \hat{x},\Delta \hat{y})\sim\text{GMM}(\bm\alpha,\bm\mu^x,\bm\mu^y,\bm\sigma^x,\bm\sigma^y,\bm\rho)$ %\Comment{Sample an offset distance}
        \STATE $\bm l_0\sim\text{Categorical}(\hat{\bm l})$ %\Comment{Sample a pen state}
        \STATE $\mathcal{\bm A}\gets\mathcal{\bm A}\cup\{[\Delta \hat{x},\Delta \hat{y}, \bm l_0]\}$ %\Comment{Save drawing action}
        \IF{$\bm l_0\neq[1,0,0]$} %\Comment{Finish stroke drawing}
            \STATE \textbf{break}
        \ENDIF
        \STATE $i\gets i+1$
    \ENDWHILE
    \STATE $\mathcal{\bm S}\gets\mathcal{\bm S}\cup\{\mathcal{\bm A}\}$ %\Comment{Save sequence-formed stroke}
    \STATE $k\gets k+1$
\ENDWHILE
\end{algorithmic}
\end{algorithm}

A pseudocode is shown in Algorithm \ref{algo1} to report the details about Sketch-HARP's generating process. A sketch is drawn stroke-by-stroke, and the generation of each stroke is divided into three stages: predicting a stroke embedding $\hat{\bm e}_1$, determining its starting position $(\hat{x},\hat{y})$ and storing it in $\mathcal{\bm P}$, and translating $\hat{\bm e}_1$ into a sequence-formed stroke $\mathcal{\bm A}$. Finally, all $\mathcal{\bm A}$s are collected by $\mathcal{\bm S}$ as the output.  

Note that error accumulation via the auto-regressive process by Sketch-HARP is slight, since each decoder hardly generates long sequences. In our hierarchical process, generating a sketch is divided into: 1) producing a sequence of stroke embeddings with positions by stroke and position decoders (stroke level), and 2) translating each embedding into a sequence of drawing actions individually (drawing action level). The lengths of sequence generated by stroke and position decoders are the number of strokes per sketch, which is around 7 for sketches in Quickdraw \cite{ha2017neural}. The length for sequence decoder is the number of drawing actions per stroke, which is around 10.

\subsection{Training a Sketch-HARP}

Our training objective is computed via sketch reconstruction, and it carries five terms as follows:

We adopt the sketch reconstruction loss from \cite{ha2017neural}, which measures the drawing-action-level differences between the generated sketch and the input, shown as $\mathcal{L}_{\text{seq}}$.
\begin{align}
	\label{eq:sequence_loss}
    &\mathcal{L}_{\text{seq}} = -\sum_{k=1}^K\sum_{i=1}^{N_k}\log \sum_{m=1}^M\alpha_{ikm}\cdot \beta_{ikm}-\sum_{k=1}^K\sum_{i=1}^{N_k}\bm l_{ki}\log\hat{\bm l}_{ki},\\
    &\beta_{ikm}=\mathcal{N}\left (\Delta x_{ki}, \Delta y_{ki}|\mu^x_{kim}, \mu^y_{kim}, \sigma^x_{kim}, \sigma^y_{kim}, \rho_{kim}\right ),
\end{align}
where $N_k$ denotes the number of drawing actions in the $k$-th stroke. $(\Delta x_{ki},\Delta y_{ki})$ is the ground truth offset distance on the canvas between the $(i-1)$-th and the $i$-th drawing actions in the $k$-th stroke. $\sum_{m=1}^M\alpha_{ikm}\cdot\beta_{ikm}$ is a P.D.F. modeled by sequence decoder in Eq.(\ref{eq:predict_action}).

The first term in $\mathcal{L}_{\text{seq}}$ is a negative log-likelihood, describing a probability that the ground truth pen positions would be sampled from our modelled mixture distribution. A small value of the first term indicates that the predicted pen positions are close to the ones in ground truth. The second term is a cross entropy, measuring whether the predicted pen states $\hat{\bm l}$ is consistent with its corresponding ground truth $\bm l$.

Furthermore, we introduce $\mathcal{L}_{\text{pos}}$ to anchor the generated strokes at their corresponding 2D coordinates on the canvas.
\begin{align}
	\label{eq:position_loss}
    \mathcal{L}_{\text{pos}} = -\sum_{k=1}^K\log\mathcal{N}\left (x_{k}, y_{k}|\mu^{px}_{k}, \mu^{py}_{k}, \sigma^{px}_{k}, \sigma^{py}_{k}, \rho^p_{k}\right ).
\end{align}
Here we also adopt the measurement in $\mathcal{L}_{\text{seq}}$ to describe whether the modeled P.D.F. could predict starting positions $\{(x_k, y_k)\}_{k=1}^K$ accurately.

As stroke decoder generates a sketch stroke-by-stroke, it is necessary to produce a marker to report when the last stroke has been drawn. For an input sketch, we utilize a sequence of vectors $\{\bm \eta_k\}_{k=1}^K$ as markers. Each $\bm \eta_k$ is a two-dimensional one-hot vector. $\bm\eta_k=[1, 0]$ or $[0, 1]$ indicate that the $k$-th stroke exists or not, respectively. We calculate the cross entropy between ground truth $\{\bm \eta_k\}$ and their corresponding predictions $\{\hat{\bm\eta}_k\}$, ensuring no redundant strokes would be drawn on the canvas.
\begin{align}
	\label{eq:stopper_loss}
    \mathcal{L}_{\text{stp}} = -\sum_{k=1}^K\bm \eta_{k}\log\hat{\bm \eta}_{k}.
\end{align}

$\mathcal{L}_{\text{seq}}$ drives the generated strokes to be consistent with the original input in drawing action level. We also introduce $\mathcal{L}_{\text{sok}}$ to assist stroke reconstruction in feature level.
\begin{align}
	\label{eq:stroke_loss}
    \mathcal{L}_{\text{sok}} = \sum_{k=1}^K\Vert\hat{\bm e}_{k}-\text{sg}(\tilde{\bm e}_{k})\Vert^2_2,
\end{align}
where $\text{sg}(\cdot)$ denotes stop gradient operation, and $\Vert\cdot\Vert_2^2$ represents L2 norm function. $\mathcal{L}_{\text{sok}}$ is a regularization term to push the generated stroke embedding $\hat{\bm e}_k$ towards its target $\tilde{\bm e}_k$.

Moreover, we adopt a convolutional neural network (CNN) decoder $p_{\text{img}}(\bm I|\bm y)$ to reconstruct the image-formed sketch $\bm I\in\mathbb{R}^{128\times 128\times 1}$ by its code $\bm y$. The reconstructed sketch image $\hat{\bm I}$ is sent into a regularization loss $\mathcal{L}_{\text{img}}$ to encourage $\bm y$ to carry more features from the input sketch.
\begin{align}
	\label{eq:image_loss}
    \mathcal{L}_{\text{img}} = \Vert\hat{\bm I}-\bm I\Vert^2_2.
\end{align}

Finally, our objective is to minimize
\begin{align}
	\label{eq:total_loss}
    \mathcal{L} &= \mathcal{L}_{\text{seq}}+\mathcal{L}_{\text{pos}}+\mathcal{L}_{\text{stp}}+5\cdot\mathcal{L}_{\text{sok}}+0.5\cdot\mathcal{L}_{\text{img}}.
\end{align}
We give $\mathcal{L}_{\text{sok}}$ a large weight to predict accurate stroke embeddings $\hat{\bm e}_k$ to further improve sketch generation.

%%%%%%%%%%%%%%%%%%%%%
\section{Experiments and Analysis}
%%%%%%%%%%%%%%%%%%%%%

\subsection{Preparations}

\noindent\textbf{Datasets}. Two datasets from QuickDraw \citep{ha2017neural} are selected. Dataset 1 (DS1), adopted from \cite{li2024sketchedit,su2020sketchhealer}, contains 17 categories (airplane, angel, alarm clock, apple, butterfly, belt, bus, cake, cat, clock, eye, fish, pig, sheep, spider, umbrella, the Great Wall of China). The sketches in DS1 are easily to be recognized in category. Dataset 2 (DS2), adopted from \cite{zang2021controllable,li2024sketchedit}, collects 5 categories (bee, bus, flower, giraffe and pig). DS2 could evaluate whether a method is powerful on learning sketch representation, as some categories contain multiple non-categorical features (e.g., giraffes orienting left or right). Each category contains $70K$ training, $2.5K$ valid and $2.5K$ test sketches ($1K=1000$).

\noindent\textbf{Baselines}. We compare Sketch-HARP with 7 baseline models: RPCL-pix2seq \cite{zang2021controllable}, RPCL-pix2seqH \cite{zang2024self}, SketchHealer \cite{su2020sketchhealer}, Lmser-pix2seq \cite{li2024lmser}, SP-gra2seq \cite{zang2023linking}, DC-gra2seq \cite{zang2025equipping} and SketchEdit \cite{li2024sketchedit}. Note that SketchEdit \cite{li2024sketchedit}, which is designed for sketch edit, is selected to make a direct comparison with Sketch-HARP on sketch manipulation.

\noindent\textbf{Implementation details}. The LSTM-based $q_{\text{sok}}$ and $q_{\text{skc}}$ are with hidden state size as 512, and $p_{\text{sok}}$, $p_{\text{pos}}$ and $p_{\text{seq}}$ are set at 1024. $q_{\text{rel}}$ is a stack of two gMLP blocks with $d_{\text{model}}=128$ and $d_{\text{ffn}}=512$. And $p_{\text{img}}$ consists of a fully connected layer (with output dimension at 2048, which are reshaped to $128\times 4\times 4$ before fed into CNN layers) and five convolutional layers (with channel numbers as 128, 128, 128, 128 and 1, kernel size $4\times 4$, stride 2 and padding 1) followed by batch normalization and ReLU activation function (the last activation function is Tanh). When training a Sketch-HARP, the mixture number $M$ in GMM distribution, the max number of strokes $N_{\text{max}}^{\text{num}}$, the max length of a stroke $N_{\text{max}}^{\text{len}}$ and the mini-batch size are fixed at 20, 25, 32, 128, respectively.

\subsection{Stroke-Level Sketch Manipulation} 

\begin{figure}[!t]
    \centering
    \includegraphics[width=0.95\columnwidth]{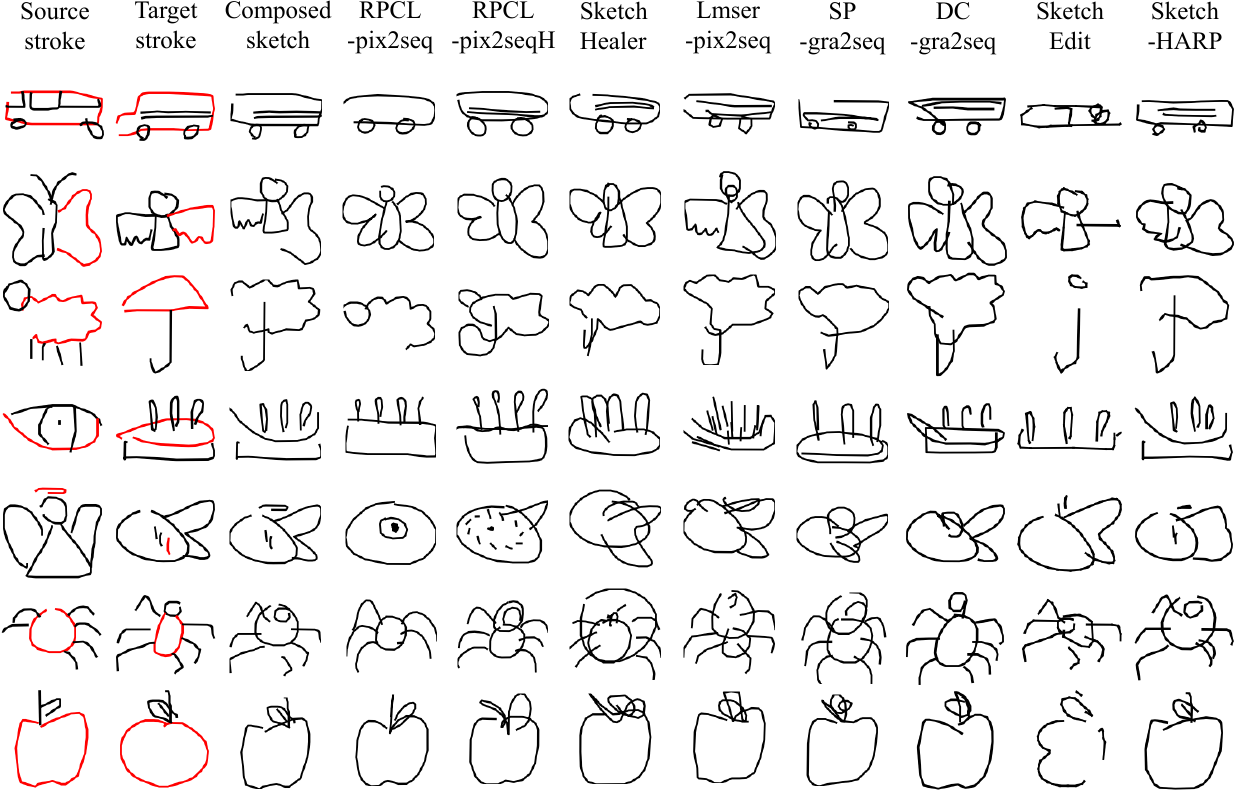}
    \caption{Comparisons on stroke replacement. For each sketch in ``target stroke'' column, its stroke in red is replaced by the highlighted one in ``source stroke'' column, with the composed sketches listed in ``composed sketch'' column.}
    \label{fig:replacement}
\end{figure}

\noindent\textbf{Stroke replacement}. This application aims to replace a target stroke from a sketch with a source stroke, which could be selected from another sketch given. For example, in Fig. \ref{fig:replacement}, the red stroke in ``target stroke'' column is replaced by the red one in ``source stroke'' column, and models are required to generate sketches to preserve details from the composed sketches. A simple way to realize it is directly replacing strokes on sketches, e.g., Fig. 3 in \cite{li2024sketchedit}, and further feed the manipulated sketch into models for feature extraction and then sketch generation. Sketch-HARP enables stroke replacement during sketch drawing by replacing a predicted target stroke embedding with another source embedding. More specifically, we could replace the predicted stroke embedding $\hat{\bm e}_1$ at \emph{Line 3} in Algorithm \ref{algo1} with another embedding $\bm\epsilon$ captured from an expected stroke, and use $\bm\epsilon$ to proceed the following generating process. Its starting position and drawing actions of $\bm\epsilon$ would be automatically generated at \emph{Line 9} and \emph{Line 24}, respectively.

Fig. \ref{fig:replacement} shows some stroke replacement results. Sketch-HARP is powerful to maintain the features from source stroke in generated sketches, though no visual patterns are utilized in sketch code learning, compared with some baselines (e.g., Lmser-pix2seq and DC-gra2seq) whose model inputs are sketch images. Furthermore, our generated sketches are harmonious in appearance, even if the introduced source stroke and the target sketch are collected from different categories. For example, in the 2-nd row, the pair of wings of the generated angel are different in shape, but are about the same size. These experimental results demonstrate that Sketch-HARP could restore stroke embeddings into drawn strokes well, and is able to customize strokes to produce reasonable sketches.

\noindent\textbf{Stroke erasion}. This application erases stroke(s) from a sketch, and the adjusted sketch is fed into models as a condition for sketch generation. Models are required to preserve the details from the adjusted sketches. Sketch-HARP could also extend stroke erasion in sketch generating process by erasing stroke embeddings instead of strokes. More specifically, if we aim to erase the currently predicted stroke embedding, it is flexible to skip position determination and stroke translation, i.e., stop recording the drawing actions of the current stroke, but continuing predicting the next stroke embedding for generation. In practice, it is flexible to skip \emph{Line 8-24} to stop calculating and storing its coordinate and drawing actions.

\begin{figure}[!t]
    \centering
    \includegraphics[width=0.95\columnwidth]{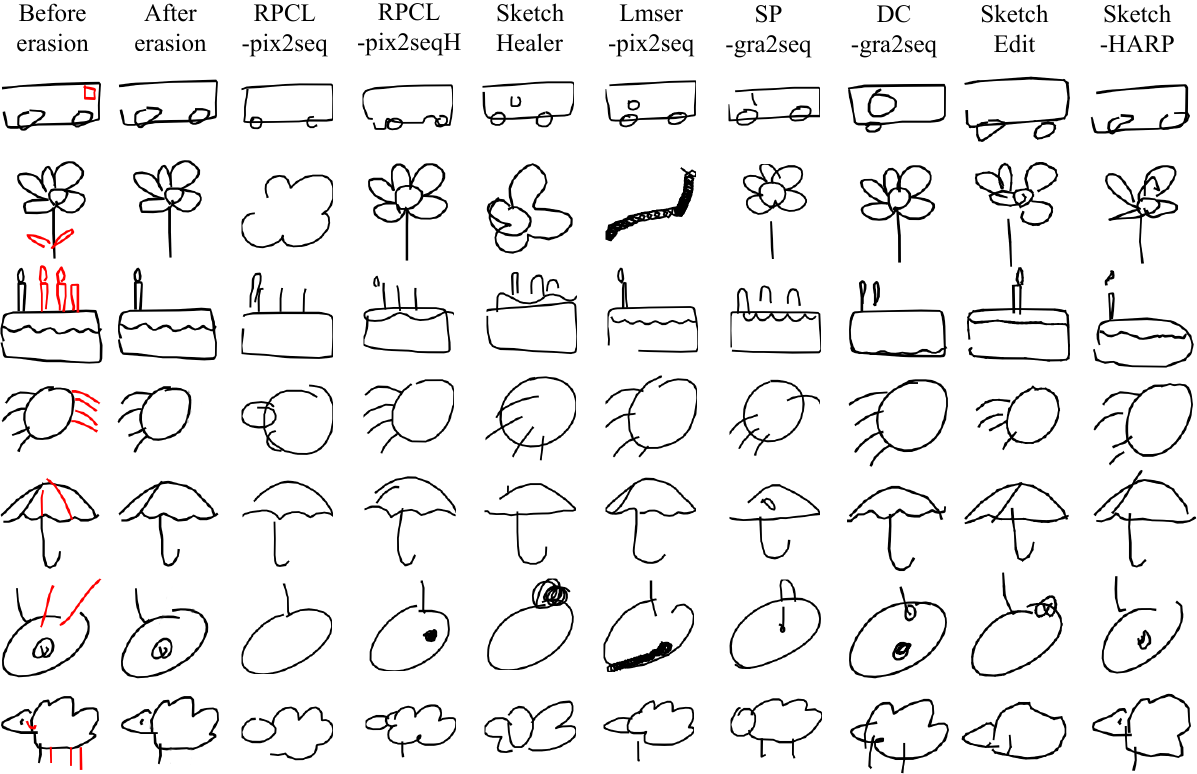}
    \caption{Comparisons on stroke erasion.}
    \label{fig:erasion}
\end{figure}

Fig. \ref{fig:erasion} shows some stroke erasion results. Sketch-HARP could prevent drawing redundant strokes when the currently generated sketch contains enough details about the input sketch. For example, only a single leg is drawn in the sheep, though it may not be common to represent a sheep. Furthermore, Sketch-HARP could always anchor strokes at their right positions, e.g., the bus wheels are closely connected with its body, and only the leftmost candle remains on the cake. These experimental results demonstrate that Sketch-HARP could precisely manipulate stroke drawing in determining strokes' positions and controlling an appropriate number of strokes.

\begin{figure}[!t]
    \centering
    \includegraphics[width=0.95\columnwidth]{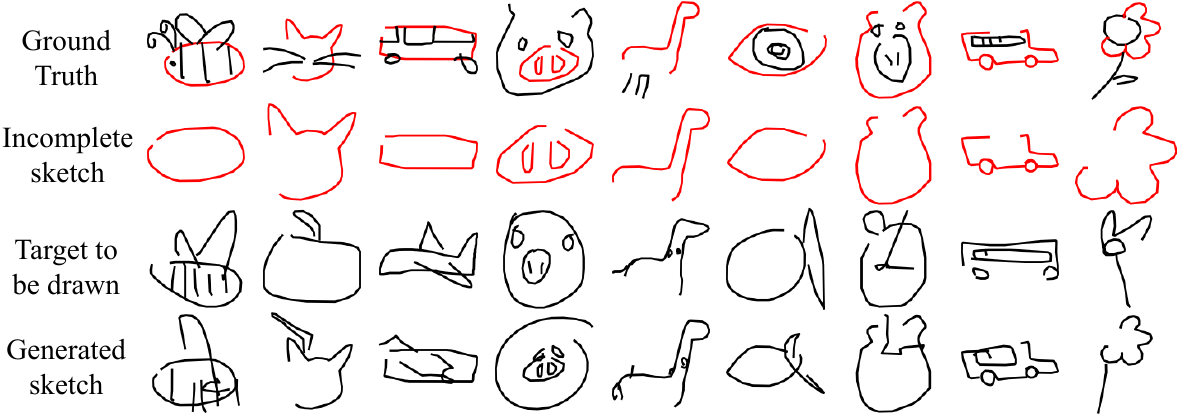}
    \caption{Exemplary stroke expansion results by Sketch-HARP. Sketch-HARP is required to continue drawing from an incomplete sketch, and the generated sketch should contain features from another source sketch.}
    \label{fig:expansion}
\end{figure}

\noindent\textbf{Stroke expansion}. Stroke expansion aims to insert some introduced strokes during sketch drawing. To realize this task, the stroke, which is expected to be inserted into a sketch, is firstly sent into stroke encoder to capture its embedding $\bm\epsilon$. During sketch drawing, at an expected time, we stop the generating process at \emph{Line 3} in Algorithm \ref{algo1} and set a breakpoint. $\bm\epsilon$ is injected into the procedure to yield its starting position $(x_{\bm\epsilon},y_{\bm\epsilon})$ and drawing strokes $\mathcal{\bm A}_{\bm\epsilon}$, which are stored at \emph{Line 18} and \emph{Line 24}, respectively. After that, we resume from the breakpoint at \emph{Line 3}, but replace $\hat{\bm e}_0$ and $\hat{\bm p}$ at \emph{Line 3} with the corresponding stroke and position embeddings $\bm\epsilon$ and $\bm p_{\bm\epsilon}$ captured from the injected stroke recently. The drawing process continues until the drawing completes. 

A special case of stroke expansion is that the introduced strokes are inserted at the start of drawing. For example, in Fig. \ref{fig:expansion}, Sketch-HARP is required to continue drawing from the unfinished sketches in 2-nd row, and patterns of the generated sketches are conditional on another references (sketches in 3-rd row). 

Fig. \ref{fig:expansion} reports some stroke expansion results by Sketch-HARP. Without visual features extracted, Sketch-HARP could still generate reasonable sketches to not only fit the categorical patterns from source sketches but also preserve detailed patterns from incomplete sketches. For example, in the 7-th column, Sketch-HARP could draw hands of a clock on a face contour borrowed from a pig, to create a pig-head-shaped clock. The features extracted from the drawn strokes could be referenced by Sketch-HARP to guide sketch drawing for creating unique hybrid sketches.

\noindent\textbf{Manipulating sketch drawing at stroke-level}. As Sketch-HARP generates sketches in a hierarchical process with exposed stroke embeddings, it is possible to edit stroke embeddings during sketch drawing to manipulate sketch drawing at stroke-level. More specifically, during sketch drawing, two basic operations: 1) erasing the stroke which was drawn recently, and 2) inserting a given stroke into generation before the current stroke drawing, could be applied to manipulate sketch drawing. And it is able to realize stroke replacement by applying stroke erasion and expansion in a queue.

\begin{figure}[!t]
    \centering
    \includegraphics[width=0.95\columnwidth]{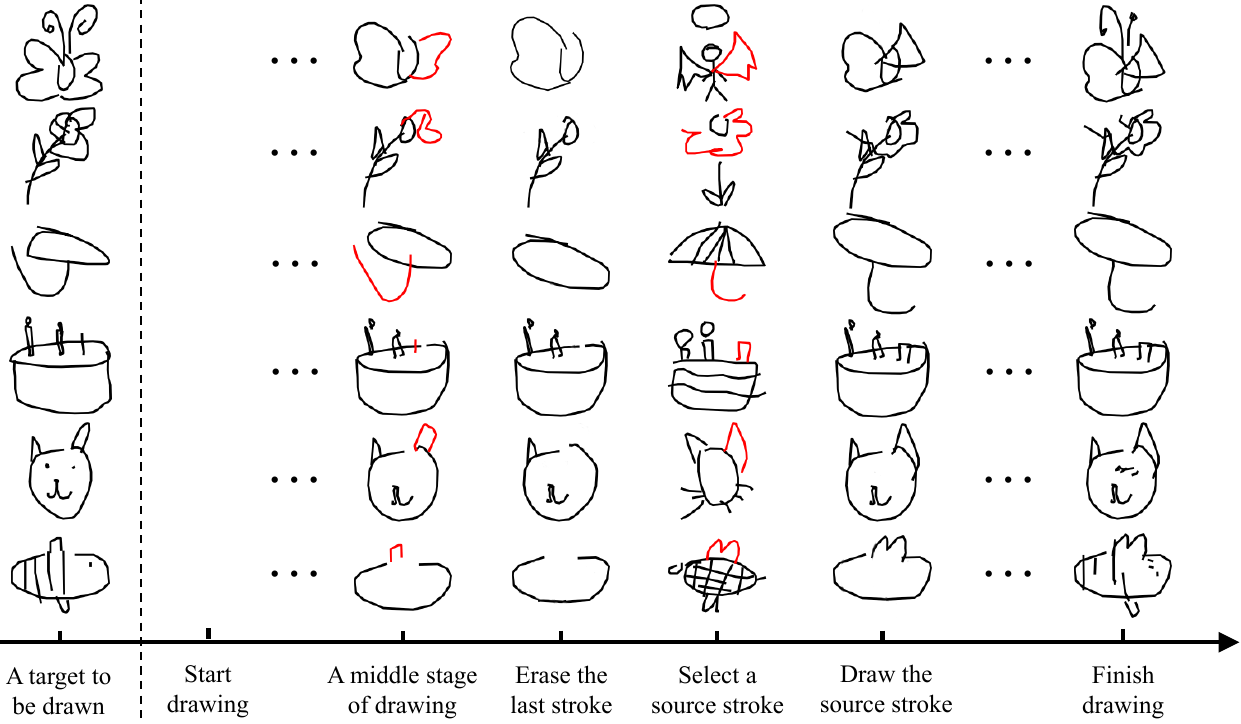}
    \caption{Manipulating sketch drawing at stroke-level by Sketch-HARP. Before applying stroke erasion, the last-drawn-stroke for each sketch is highlighted in red.}
    \label{fig:manipulate}
\end{figure}

Fig. \ref{fig:manipulate} reports some results by Sketch-HARP. When a sketch is being drawn at a middle stage, it is free to erase the last-drawn-stroke (highlighted in red in the 2-nd column). Moreover, we could select a source stroke from another sketch and inject its stroke embedding into Sketch-HARP's generator to draw the source stroke as a stroke expansion. After that, Sketch-HARP is released to finish sketch drawing freely. As shown in Fig. \ref{fig:manipulate}, our model could preserve both features from the target sketch and from the selected stroke to yield recognizable sketches after manipulation.

\noindent\textbf{Human perception quality}. In order to investigate human subjective rating of the quality of sketch manipulation, we administer a questionnaire with 10 sketch replacement and 10 sketch erasion questions. In the questions for stroke replacement, for a full sketch, some target strokes colored in red would be replaced by the source stroke(s) collected from another sketch, obtaining the manipulated sketch. And in the questions for stroke erasion, some strokes in red are erased from the given sketch to get the manipulated sketch. Each manipulated sketch is fed into 8 models respectively as a reference, and models are required to generate sketches to preserve detailed patterns from the given input. In a question, human volunteers are asked to select and rank the best three generated sketches.

\begin{table}[!t]
    \centering
    \small
    \begin{tabular}{@{}lcc@{}}
        \hline
        Model & Replacement & Erasion\\
        \hline
		RPCL-pix2seq & 0.21$\pm$0.59 & 0.38$\pm$0.76\\
		RPCL-pix2seqH & 0.29$\pm$0.71 & 0.46$\pm$0.76\\
		SketchHealer & 0.36$\pm$0.73 & 0.53$\pm$0.94 \\
		Lmser-pix2seq & 0.64$\pm$0.93 & 0.65$\pm$1.02\\
		SP-gra2seq & 0.52$\pm$0.79 & 0.40$\pm$0.78 \\
		DC-gra2seq & 0.44$\pm$0.90 & 0.24$\pm$0.65\\
        SketchEdit & 1.45$\pm$1.27 & 1.04$\pm$1.14 \\
        Sketch-HARP & \pmb{2.09$\pm$1.06} & \pmb{2.30$\pm$0.96}\\
        \hline
    \end{tabular}
    \caption{Human scoring results (mean$\pm$std.) on stroke replacement and stroke erasion.}
    \label{tab:human_scoring}
\end{table}

For each question, a human subject was required to rank the manipulated sketches generated by Sketch-HARP and 7 baselines, and we only value the best three models with scores at 3, 2 and 1, respectively, leaving the rest models with 0 score. 51 volunteers participated in the evaluation, and the scoring results in Table \ref{tab:human_scoring} reports that Sketch-HARP outperforms all baselines. We also raise t-tests in which the p-values against our Sketch-HARP with SketchEdit are $6.04\times10^{-13}$ and $1.23\times10^{-56}$ for sketch replacement and erasion, respectively, revealing that the improvement contributed by Sketch-HARP compared with SketchEdit is significant.

Details about the questionnaire, the scoring rules and the human perception evaluation are listed below.

\begin{figure*}[!t]
    \centering
    \includegraphics[width=1.8\columnwidth]{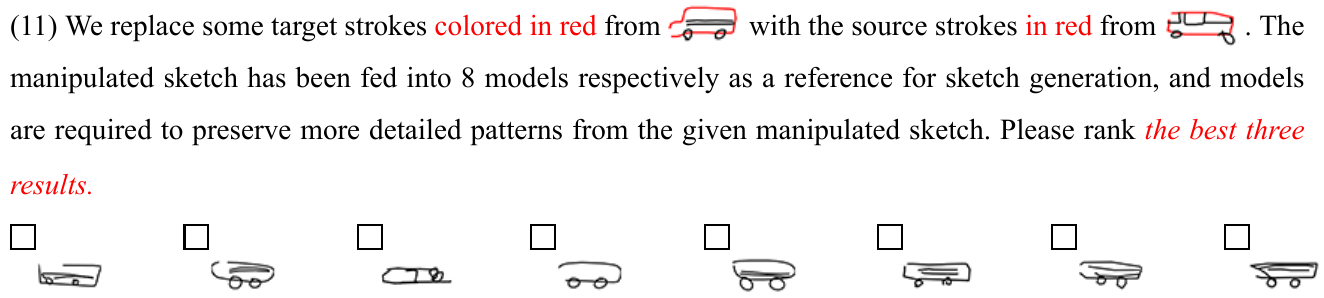}
    \caption{An exemplary question for stroke replacement.}
    \label{fig:questionnaire_replacement}
\end{figure*}

\begin{figure*}[!t]
    \centering
    \includegraphics[width=1.8\columnwidth]{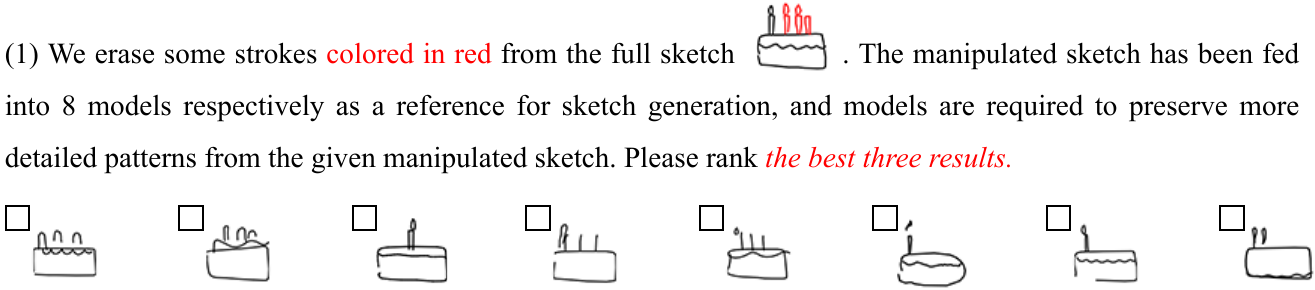}
    \caption{An exemplary question for stroke erasion.}
    \label{fig:questionnaire_erasion}
\end{figure*}

\emph{Questionnaire details}. The questionnaire consists of 20 questions, with 10 sketch replacement and 10 sketch erasion questions, respectively. Fig. \ref{fig:questionnaire_replacement} and Fig. \ref{fig:questionnaire_erasion} show two exemplary questions for each type, respectively. 

\emph{Scoring rules and human perception evaluation}. For the 8 generated sketches (by Sketch-HARP and 7 baseline models) in each question, the best three are selected and ranked by each human subject. We offer the corresponding best three models with scores 3, 2 and 1, respectively, leaving the rest models with 0 scores. As totally 51 volunteers have participated in this evaluation and there are 10 questions per type (stroke replacement and erasion), we collected $51\times 10=510$ scores for each question type and computed their mean values and standard deviations, respectively.

\subsection{Sketch Representation and Reconstruction}

We also make comparisons on sketch representation and reconstruction. If the generated sketches preserve more details from the inputs, that model is powerful in both sketch representation learning and conditional sketch generation. We utilize the Fr\'{e}chet Inception Distance (FID) score \cite{heusel2017gans}, the learned perceptual image patch similarity (LPIPS) computed by ControlNet \cite{zhang2023adding} and the contrastive language-image pre-training score (CLIP-S) \cite{hessel2021clipscore} to evaluate the generative performance on sketches. We also use $Rec$ and $Ret$ \cite{zang2021controllable} to measure whether the generated sketch could preserve more categorical or stylistic features from the input, and we trained two sketch-a-nets \cite{yu2015sketch} on DS1 and DS2, respectively, for computing $Rec$. Since SketchEdit represents a sketch by a group of stroke embeddings instead of a single sketch code, its generator receives about 10 times of information from the input sketch than all the other models. To make the comparison fair, we do not include it in this quantitative comparison.

\begin{table*}[!ht]
    \centering
    \small
    \begin{tabular}{@{}clrcccrrr@{}}
        \hline
        \multirow{2}*{Dataset} & \multirow{2}*{Model} & \multirow{2}*{FID~$\downarrow$} & \multirow{2}*{LPIPS~$\downarrow$} & \multirow{2}*{CLIP-S~$\uparrow$} & \multirow{2}*{$Rec\uparrow$} & \multicolumn{3}{c}{$Ret\uparrow$} \\ \cline{7-9}
		~ & ~ & ~ & ~ & ~ & ~ & @1 & @10 & @50 \\
        \hline
		\multirow{8}*{DS1} & RPCL-pix2seq \cite{zang2021controllable} & 22.74 & 0.37 & 88.57 & 81.80 & 28.80 & 59.05 & 77.52 \\
		~ & RPCL-pix2seqH \cite{zang2024self} & 17.30 & 0.38 & 88.66 & 87.82 & 81.22 & 92.15 & 94.90\\
		~ & SketchHealer \cite{su2020sketchhealer} & 27.72 & 0.39 & 88.17 & 87.03 & 68.52 & 82.37 & 86.57 \\
		~ & Lmser-pix2seq \cite{li2024lmser} & 18.94 & 0.36 & 89.74 & \pmb{90.50} & 80.56 & 86.68 & 89.88\\
		~ & SP-gra2seq \cite{zang2023linking} & 10.80 & 0.38 & 88.72 & 89.83 & 94.05 & 98.72 & 99.57 \\
		~ & DC-gra2seq \cite{zang2025equipping} & 12.83 & 0.30 & \pmb{93.84} & 90.41 & 96.27 & \pmb{99.47} & \pmb{99.81} \\
        ~ & Sketch-HARP & \pmb{9.96} & \pmb{0.28} & 91.60 & 89.90 & \pmb{97.96} & 99.42 & 99.77\\
        \hline
        \multirow{8}*{DS2} & RPCL-pix2seq \cite{zang2021controllable} & 36.56 & 0.28 & 85.47 & 74.94 & 26.72 & 48.80 & 63.13 \\
		~ & RPCL-pix2seqH \cite{zang2024self} & 26.90 & 0.34 & 88.13 & 84.40 & 74.63 & 90.09 & 94.65\\
		~ & SketchHealer \cite{su2020sketchhealer} & 31.02 & 0.29 & 91.46 & 74.93 & 32.09 & 57.35 & 73.63 \\
		~ & Lmser-pix2seq \cite{li2024lmser} & 10.10 & 0.32 & 91.10 & 85.02 & 90.23 & 93.05 & 94.67\\
		~ & SP-gra2seq \cite{zang2023linking} & 34.06 & 0.45 & 85.17 & 76.89 & 56.27 & 80.26 & 90.56 \\
		~ & DC-gra2seq \cite{zang2025equipping} & 11.01 & 0.30 & \pmb{94.46} & 85.67 & 95.27 & 99.09 & 99.65\\
        ~ & Sketch-HARP & \pmb{6.97} & \pmb{0.22} & 93.31 & \pmb{87.71} & \pmb{96.00} & \pmb{99.78} & \pmb{99.93}\\
        \hline
    \end{tabular}
    \caption{Sketch reconstruction performance (\%). ``@$k$'' indicates the top-$k$ accuracy.}
    \label{tab:controllable}
\end{table*}

\begin{figure}[!t]
    \centering
    \includegraphics[width=0.95\columnwidth]{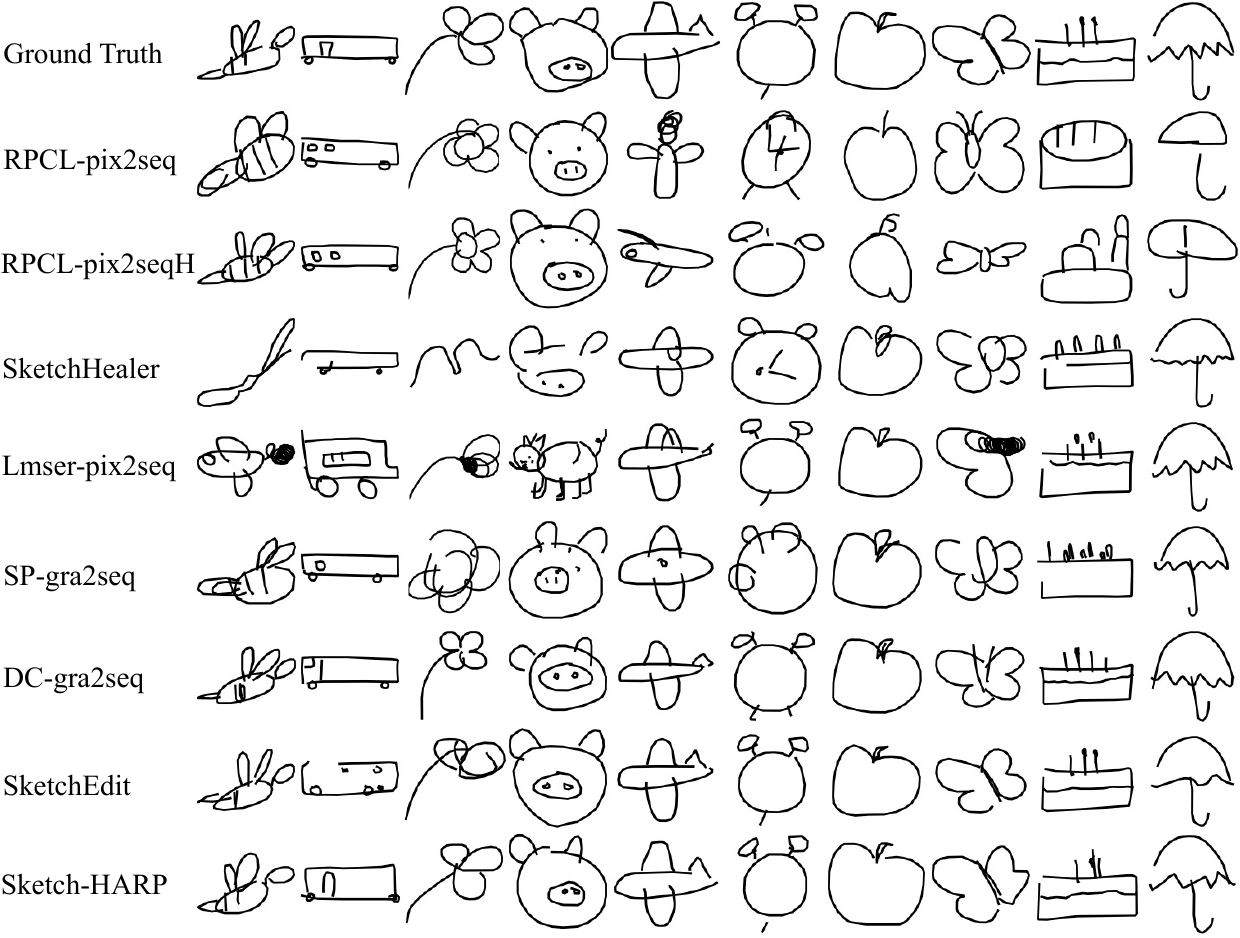}
    \caption{Qualitative comparisons on sketch reconstruction.}
    \label{fig:reconstruction}
\end{figure}

Table \ref{tab:controllable} reports quantitative results on two datasets. Sketch-HARP achieves comparable sketch reconstruction performance with state-of-the-art baselines. DC-gra2seq, which is the state-of-the-art model on sketch reconstruction, learns sketch codes from sketch images cooperated with contextual information captured from drawing orders. The visual patterns of sketches, which are never utilized by Sketch-HARP, improves DC-gra2seq's reconstruction performance. Besides, Fig. \ref{fig:reconstruction} reports qualitative comparisons on sketch reconstruction.

\subsection{Ablation Studies} 

\noindent\textbf{Performance gained from relationship embeddings}. We discuss whether the learned relationship embeddings $\bm r_k$ among strokes in Eq. (\ref{eq:relationship_embedding}) are beneficial on sketch generation. We train a variant of Sketch-HARP without learning relationship embeddings, and Table \ref{tab:ablation_1} reports its performance on sketch reconstruction. 

\begin{table*}[!t]
    \centering
    \small
    \begin{tabular}{@{}ccccccrrr@{}}
        \hline
    \multirow{2}*{Dataset} & \multirow{2}*{\makecell{Relationship\\ embeddings}} & \multirow{2}*{FID~$\downarrow$} & \multirow{2}*{LPIPS~$\downarrow$} & \multirow{2}*{CLIP-S~$\uparrow$} & \multirow{2}*{$Rec\uparrow$} & \multicolumn{3}{c}{$Ret\uparrow$} \\ \cline{7-9}
		~ & ~ & ~ & ~ & ~ & ~ & @1 & @10 & @50 \\
        \hline
		\multirow{2}*{DS1} & \checkmark & \pmb{9.96} & \pmb{0.28} & \pmb{91.60} & \pmb{89.90} & \pmb{97.96} & \pmb{99.42} & \pmb{99.77}\\
        ~ & $\times$ & 25.45 & 0.36 & 87.69 & 74.55 & 95.43 & 98.58 & 99.58\\
        \hline
        \multirow{2}*{DS2} & \checkmark & \pmb{6.97} & \pmb{0.22} & \pmb{93.31} & \pmb{87.71} & \pmb{96.00} & \pmb{99.78} & \pmb{99.93}\\
        ~ & $\times$ & 25.95 & 0.34 & 88.93 & 61.79 & 92.45 & 96.19 & 97.71\\
        \hline
    \end{tabular}
    \caption{Learning relationship embeddings is beneficial to sketch reconstruction.}
    \label{tab:ablation_1}
\end{table*}

\begin{figure*}[!t]
    \centering
    \includegraphics[width=1.8\columnwidth]{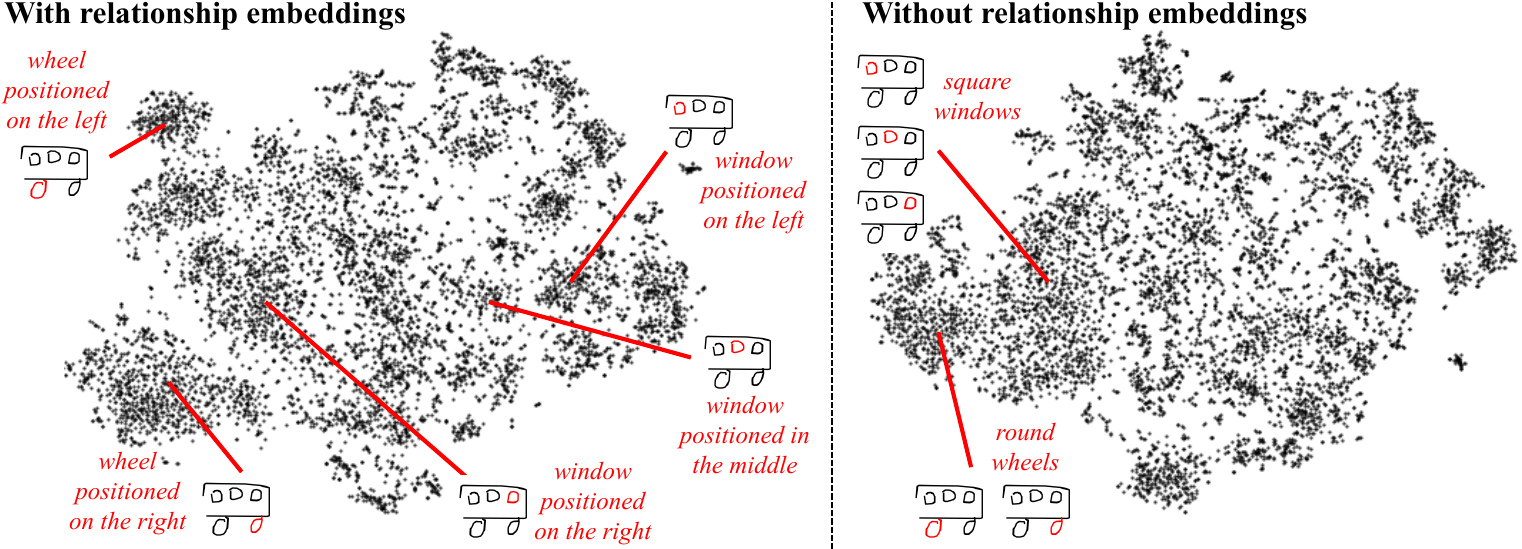}
    \caption{Comparisons between latent spaces of $\tilde{\bm e}$ (with relationship embeddings) and $\bm e$ (without relationship embeddings) for category bus.}
    \label{fig:ablation}
\end{figure*}

Without receiving the captured spatial, contextual and semantic relationships among stroke from relationship embeddings, a stroke embedding has no chance to sense the entire sketch in a global view. Thus, the sketch reconstruction performance reduces with the absence of relationship embeddings. Fig. \ref{fig:ablation} also reveals the latent spaces of $\tilde{\bm e}$ (with relationship embeddings) and $\bm e$ (without relationship embedding) for category bus via $t$-SNE \cite{maaten2008visualizing}. With the attendance of relationship embeddings, stroke embeddings are organized into compact clusters according to their features. Thus, wheels positioned on the left or right, windows located in different positions are projected into unique clusters, driving Sketch-HARP to predict appropriate stroke embeddings accurately in sketch drawing to further benefit sketch generation.

\noindent\textbf{The impact of sketch image reconstruction}. We discuss whether the introduced branch of sketch image reconstruction along with the corresponding loss term $\mathcal{L}_{\text{img}}$ in Eq. (\ref{eq:image_loss}) benefit sketch generation. A variant of Sketch-HARP by removing the loss term of $\mathcal{L}_{\text{img}}$ is trained, and the results are shown in Table \ref{tab:ablation_2}.

Without regularizing latent codes of sketches via sketch image reconstruction, a sketch code $\bm y$ may fail to represent visual patterns in a global view from an entire sketch. Though the model could still focus on generating well-drawn sketch sequences at stroke level with the rest loss terms, the visual appearance of the generated sketches may be unrecognizable sometimes, which results in the poor performance on $Rec$ in Table \ref{tab:ablation_2}.

\begin{table*}[!t]
    \centering
    \small
    \begin{tabular}{@{}ccccccrrr@{}}
        \hline
        \multirow{2}*{Dataset} & \multirow{2}*{\makecell{Reconstruct\\ sketch image}} & \multirow{2}*{FID~$\downarrow$} & \multirow{2}*{LPIPS~$\downarrow$} & \multirow{2}*{CLIP-S~$\uparrow$} & \multirow{2}*{$Rec\uparrow$} & \multicolumn{3}{c}{$Ret\uparrow$} \\ \cline{7-9}
		~ & ~ & ~ & ~ & ~ & ~ & @1 & @10 & @50 \\
        \hline
		\multirow{2}*{DS1} & \checkmark & \pmb{9.96} & \pmb{0.28} & \pmb{91.60} & \pmb{89.90} & \pmb{97.96} & \pmb{99.42} & 99.77\\
        ~ & $\times$ & 9.98 & 0.28 & 91.48 & 70.18 & 97.79 & 99.28 & \pmb{99.78}\\
        \hline
        \multirow{2}*{DS2} & \checkmark & \pmb{6.97} & \pmb{0.22} & \pmb{93.31} & \pmb{87.71} & \pmb{96.00} & \pmb{99.78} & \pmb{99.93}\\
        ~ & $\times$ & 7.41 & 0.24 & 92.66 & 58.07 & 95.82 & 99.36 & 99.72\\
        \hline
    \end{tabular}
    \caption{The impact of sketch image reconstruction loss term $\mathcal{L}_{\text{img}}$ in Eq. (\ref{eq:total_loss}) on sketch reconstruction.}
    \label{tab:ablation_2}
\end{table*}

\begin{table*}[!t]
    \centering
    \small
    \begin{tabular}{@{}ccrcccrrr@{}}
        \hline
        \multirow{2}*{Dataset} & \multirow{2}*{\makecell{Stroke embedding\\ regularization}} & \multirow{2}*{FID~$\downarrow$} & \multirow{2}*{LPIPS~$\downarrow$} & \multirow{2}*{CLIP-S~$\uparrow$} & \multirow{2}*{$Rec\uparrow$} & \multicolumn{3}{c}{$Ret\uparrow$} \\ \cline{7-9}
		~ & ~ & ~ & ~ & ~ & ~ & @1 & @10 & @50 \\
        \hline
		\multirow{2}*{DS1} & \checkmark & \pmb{9.96} & \pmb{0.28} & \pmb{91.60} & \pmb{89.90} & \pmb{97.96} & \pmb{99.42} & \pmb{99.77}\\
        ~ & $\times$ & 52.31 & 0.42 & 86.83 & 35.93 & 84.75 & 93.46 & 96.42\\
        \hline
        \multirow{2}*{DS2} & \checkmark & \pmb{6.97} & \pmb{0.22} & \pmb{93.31} & \pmb{87.71} & \pmb{96.00} & \pmb{99.78} & \pmb{99.93}\\
        ~ & $\times$ & 74.05 & 0.40 & 87.34 & 17.97 & 81.66 & 88.21 & 89.88 \\
        \hline
    \end{tabular}
    \caption{The impact of stroke embedding regularization loss term $\mathcal{L}_{\text{sok}}$ in Eq. (\ref{eq:total_loss}) on sketch reconstruction.}
    \label{tab:ablation_3}
\end{table*}

\noindent\textbf{The impact of regularizing stroke embeddings}. We further discuss the impact of regularizing the generated stroke embeddings $\hat{\bm e}$ with the corresponding ground truth $\tilde{\bm e}$. We train a variant of Sketch-HARP by canceling the loss term $\mathcal{L}_{\text{sok}}$ in Eq. (\ref{eq:total_loss}). Experimental results are shown in Table \ref{tab:ablation_3}.

When $\mathcal{L}_{\text{sok}}$ is removed from the objective, during sketch generation, each predicted stroke embedding by stroke decoder would not be constrained at its corresponding ground truth, i.e., the stroke embedding captured from the exact input stroke. As a result, it is much more difficult to guide Sketch-HARP to produce a group of ordered stroke (embeddings) to complete the original sketch input, leading to poor performance on sketch reconstruction.

%%%%%%%%%%%%%%%%%%%%%
\section{Conclusions}
%%%%%%%%%%%%%%%%%%%%%

We have presented Sketch-HARP, a sequence-formed sketch generative model with a three-staged hierarchical auto-regressive generating process, designed for flexibly manipulating sketch drawing at stroke-level. By separating a stroke drawing into stroke embedding prediction, stroke position determination and drawing actions translation, the stroke embeddings, which are introduced as editable accesses in generating process, are exposed for manipulation. Thus, we are able to control the strokes' features, their locations on the canvas, and the number of strokes, i.e., manipulating sketch drawing in a controllable way.

\section{Acknowledgments}
This work was supported by the National Natural Science Foundation of China (62406064) and the Fundamental Research Funds for the Central Universities (2232024D-28).

\bibliography{reference}

\end{document}